\documentclass[10pt,twocolumn,letterpaper]{article}

\usepackage{iccv}
\usepackage{times}
\usepackage{epsfig}
\usepackage{graphicx}
\usepackage{amsmath}
\usepackage{amssymb}
\usepackage{booktabs}
\usepackage{slashbox}
\usepackage{mathrsfs}  
\usepackage{eucal}
\usepackage{enumerate}
\usepackage{array}

\graphicspath{ {./} }

\usepackage[colorlinks,breaklinks=true,bookmarks=false]{hyperref}
\usepackage[all]{hypcap}
\iccvfinalcopy 

\newcommand{\PreserveBackslash}[1]{\let\temp=\\#1\let\\=\temp}
\newcolumntype{C}[1]{>{\PreserveBackslash\centering}p{#1}}
\newcolumntype{R}[1]{>{\PreserveBackslash\raggedleft}p{#1}}
\newcolumntype{L}[1]{>{\PreserveBackslash\raggedright}p{#1}}

\ificcvfinal\fi

\def\degree{${}^{\circ}$}

\begin{document}

\title{End-to-End Face Parsing via Interlinked Convolutional Neural Networks}

\author{Zi Yin$^{1}$ \ \ Valentin Yiu$^{2,3}$ \ \ Xiaolin Hu $^{2}$ \ \ Liang Tang$^{1}$\thanks{Corresponding author.} \\
$^{1}$School of Technology, Beijing Forestry University, Beijing 100083, China. \\
$^{2}$State Key Laboratory of Intelligent Technology and Systems, \\
Institute for Artificial Intelligence, Department of Computer Science and Technology, \\
Beijing National Research Center for Information Science and Technology, THBI, Tsinghua University \\
$^{3}$CentraleSupélec, Gif-Sur-Yvette 91190, France. \\
{\tt\small \{yinzi, happyliang\}@bjfu.edu.cn \ xlhu@tsinghua.edu.cn \ yaoyq18@mails.tsinghua.edu.cn}
}

\maketitle
\ificcvfinal\thispagestyle{empty}\fi

\begin{abstract}
Face parsing is an important computer vision task that requires accurate pixel segmentation of facial parts (such as eyes, nose, mouth, etc.), providing a basis for further face analysis, modification, and other applications. Interlinked Convolutional Neural Networks (iCNN) was proved to be an effective two-stage model for face parsing. However, the original iCNN was trained separately in two stages, limiting its performance. To solve this problem, we introduce a simple, end-to-end face parsing framework: STN-aided iCNN(STN-iCNN), which extends the iCNN by adding a Spatial Transformer Network (STN) between the two isolated stages. The STN-iCNN uses the STN to provide a trainable connection to the original two-stage iCNN pipeline, making end-to-end joint training possible. Moreover, as a by-product, STN also provides more precise cropped parts than the original cropper. Due to these two advantages, our approach significantly improves the accuracy of the original model. Our model achieved competitive performance on the Helen Dataset, the standard face parsing dataset. It also achieved superior performance on CelebAMask-HQ dataset, proving its good generalization. Our code has been released at \url{https://github.com/aod321/STN-iCNN}.
\end{abstract}

Face parsing is a special case of semantic image segmentation which is the task of marking each pixel on the face with its semantic part label such as left eyebrow, upper lip, etc. Compared to face alignment \cite{jin2017face}, face parsing can provide more accurate areas, which is necessary for a variety of applications such as face understanding, modification and expression recognition. 

Convolutional neural networks (CNNs) is a type of artificial neural network, which was inspired by the existence of simple cells and complex cells \cite{hubel1962receptive}. It has been widely used in many fields, such as: recognition and classification \cite{NIPS2012_4824, simonyan2014very,zeng2018eeg,abbasi2020detecting,oyedotun2017banknote}, object detection \cite{girshick2014rich,ren2015faster}, part and keypoint prediction \cite{zhang2014part,long2014convnets}, and local correspondence \cite{long2014convnets,fischer2014descriptor}. In particular, CNNs achieve the dominant performance on semantic segmentation \cite{chen2017rethinking,long2015fully} and face parsing. CNNs models applied on those tasks are divided into global-based and region-based methods.

In recent years, region-based methods, such as Mask R-CNN \cite{he2017mask}, have achieved significant performance. Unlike global methods that apply semantic segmentation on an entire image \cite{long2015fully,ronneberger2015u,badrinarayanan2017segnet,chen2018encoder,liu2019auto}, the region-based methods only needs to focus on the semantic region, so external components from the environment have no effect on the prediction. In contrast with other segmentation tasks, facial parts are closely related to each other and their dependence on external objects is limited, thus region-based segmentation is more suitable for face parsing. Some works have achieved region-based face parsing through RoI align \cite{lin2019face} or cropping \cite{luo2012hierarchical,liu2017face} , and have achieved impressive results. However, the RoI align method requires additional landmarks to achieve RoI positioning.

Zhou \etal \cite{zhou2015interlinked} proposed a region-based face parsing framework, with a new CNN structure named interlinked Convolutional Neural Network (iCNN). Unlike the previous CNN methods, iCNN used the image pyramid method to deal with multi-scale problems, and introduced recurrent bidirectional interconnections to incorporate more comprehensive features. The pipeline of iCNN can be divided into two stages: parts cropping and parts labeling. The first stage is to crop facial parts, and the second stage is to do segmentation on individual parts. This method uses rough labeling to implement location positioning, so no additional location annotations are required. 

The pipeline of iCNN divides a one-step labeling task on a large image into two-step labeling tasks on several smaller images, saving computations. However, the cropper of \cite{zhou2015interlinked} is not differentiable. Therefore, the two stages cannot be trained in an end-to-end manner. This makes the loss of the fine labeling stage unable to help optimize the coarse labeling stage.

In order to solve this problem, we introduce STN-aided iCNN (STN-iCNN) by adding a Spatial Transformer Network (STN) \cite{jaderberg2015spatial} to the original pipeline. STN is a differentiable module that performs spatial transformation on feature maps. We use STN as a differentiable position predictor and cropper so that the entire model can be trained end-to-end. End-to-end training can optimize the two-stage model towards the common goal, so that the fine labeling stage can improve the rough labeling stage, thereby improving the overall accuracy of the model.

In addition, we also observed that under the same rough mask input, STN can provide more accurate cropped parts than the original cropper even without end-to-end training. In extreme cases, we find that even if the label of the corresponding component in the rough mask is missing, STN can still crop out that part correctly. 
This is because the added deep localization network learns the relative relationship between the various parts of the face from the rough mask, so even if the information of some parts is lost, the position of the missing part can still be inferred from the context. 
This feature makes the prediction results of our method more stable and also helps to improve the overall system performance. Experiments on the HELEN \cite{smith2013exemplar} dataset show that our method greatly improves the accuracy of the original model without changing the structure of the iCNN. Our experiments on the CelebAMask-HQ dataset further proved the validity of the model and its generalization ability.

The paper is organized as follows: Section \ref{sec:related} covers related works on face parsing discussing their strengths and weaknesses, Section \ref{sec:method} describes our proposed framework and model in detail, Section \ref{sec:exp} presents the experimental results. Finally, Section \ref{sec:conc} provides a brief summary of the study and future works.

\begin{figure*}[t]
  \begin{center}
    \includegraphics[width=\linewidth]{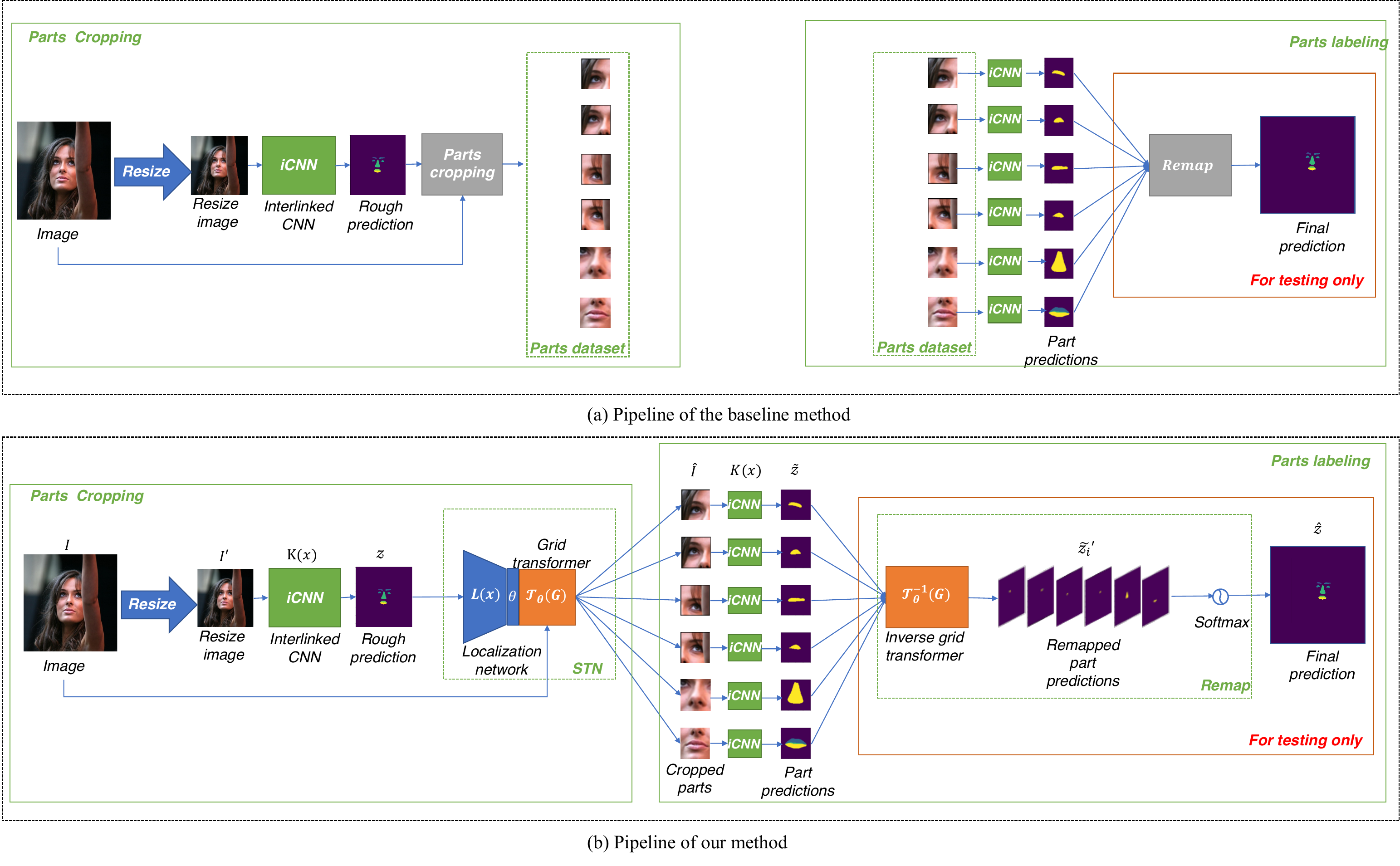}
  \end{center}
  \vspace{-1mm}
     \caption{Comparison of our method and the baseline method.}
  \label{fig:pipeline}
\end{figure*}

\section{Related Works} \label{sec:related}
\textbf{Face Parsing}.
Most existing deep learning methods for face parsing can be divided into region-based methods and global methods. 

Global methods directly perform semantic segmentation on the entire image. Early approaches include epitome model \cite{warrell2009labelfaces} and exemplar-based method \cite{smith2013exemplar}. The success of deep convolutional neural network (CNN) models has brought drastic advances in computer vision tasks \cite{NIPS2012_4824}, many face parsing methods using CNN have been proposed. Jackson \etal \cite{jackson2016cnn} used extra landmarks as guidance, using boundary cues to locate facial parts. Liu \etal \cite{liu2015multi} used a hybrid framework which contains CRF and CNN to model pixel-wise likelihoods and label dependencies jointly. Zhou \etal \cite{zhou2017face} proposed an architecture which combine fully convolutional network \cite{long2015fully}, super-pixel information, and CRF model together. Wei \etal \cite{wei2017learning} proposed a CNN network framework that can adaptively adjust the receptive fields in the middle layer and obtain better receptive fields on face parsing tasks. These models can usually be trained end-to-end. However, their performances can be improved, as the optimization cannot focus on each individual part separatly.

Region-based approaches independently predict pixel-level labels for each part by training separate models for each facial component. Luo \etal \cite{luo2012hierarchical} proposed a hierarchical structure to treat each detected facial parts separately. Liu \etal \cite{liu2017face} achieved state-of-the-art accuracy while maintaining a very fast running speed by combining a shallow CNN with a spatially variable RNN. 
In the work of Zhou \etal \cite{zhou2015interlinked}, region positioning and face parsing were accomplished with the same network structure, without the need for additional landmark detection or extra annotations. They divided the processing into two isolated stages trained independently. They designed the first stage to get a rough mask of the whole image which was used to calculate the coordinates of the facial parts. The second stage was designed to perform fine labeling for each facial parts individually. Finally, the outputs of the second stage were remapped back according to the coordinates, thereby achieving a complete face segmentation process. 

On the other hand, Lin \etal \cite{lin2019face} used a hybrid meth-od of global and region-based approaches, they solved the problem of the variable shape and size of hair by using tanh-warping and then performed segmentation globally using FCN. Moreover, similarly to Mask R-CNN \cite{he2017mask}, they used RoI-Align for region-based segmentation of the facial parts.

\textbf{General Semantic Segmentation}.
Face parsing is essentially a specific semantic segmentation for face. In recent years, general semantic segmentation or instance segmentation has achieved remarkable results. Many deep learning methods have been proposed to solve these problems. Fully convolutional neural networks (FCN) replace the last few fully connected layers to convolutional layers to enable efficient end-to-end learning and prediction \cite{long2015fully}. Based on the FCN, many other improvements, such as SegNet \cite{badrinarayanan2017segnet}, Unet \cite{ronneberger2015u} , CRFasRNN \cite{zhu2016adversarial}, DeepLab \cite{chen2017deeplab} are proposed. Compared with general semantic segmentation tasks, face parsing has only a few specific semantics (such as nose, eyes, mouth etc.), and these semantics have extremely related position and size relationships. Applying such models directly to face parsing is often incapable of utilizing these context relations \cite{lin2019face}.


\section{Method}
\label{sec:method}

\subsection{Overall Pipeline}
\label{subsec:pipeline}

The work of Zhou \etal \cite{zhou2015interlinked} was used as our baseline method. As shown in Fig. \ref{fig:pipeline}a, the baseline method is divided in two steps. The first step is to detect and crop the face parts. The second step is to label separately the cropped parts. Because the cropping method used in this process is not differentiable, these two stages cannot be jointly trained. This limits the performance of the system. 
Our proposed method solves this problem by adding a spatial transformer network (STN) between the two steps of the baseline method. STN replaces the original cropper with a differentiable spatial transformer, allowing the model to be trained end-to-end.
 
 As shown in Fig. \ref{fig:pipeline}b, for each input image, the image is first resized and passed to the iCNN model which performs coarse segmentation. Then, the predicted rough mask is sent to STN, and the localization network of the STN predicts the transformer parameter matrix $\theta$. After that, The cropped-out parts are sent to the features segmentation model. Then with $\theta$ as a parameter, the grid transformer crops corresponding parts from the original image. Finally, the inverse grid transformer remaps all the partial predictions into a final whole prediction.

Given an image $I \in \mathbb{R}^{C \times H \times W}$, it is first squarely resized into $I' \in \mathbb{R}^{C \times H' \times W'}$, where $C$ is the number of channels in the original image, $H'=W'$. $I'$ is then processed by an iCNN network $K$ which performs coarse labeling to get a rough prediction $z$:
	\begin{equation}
	 z  = K(I'),
	\end{equation}
where $z \in \mathbb{R}^{H' \times W'}$.

The rough prediction $ z $ is then processed by the localized network $ L $ as part of the STN to obtain the transformation parameter matrix $ \theta $: 
	\begin{equation}
	 \theta  = L(z),
	\end{equation}
where $\theta \in \mathbb{R}^{N \times 2 \times 3}$, $N$ is the number of individual components. Given $ \theta $, the grid transformer crops the individual parts $\hat I_i $ from the original image $I$:
	\begin{equation}
	\hat I_i  =   \mathcal{T}_{\theta}(I_i),
	\end{equation} 
where $i \in \{0, 1, .., N\}$, $\hat I_i \in \mathbb{R}^{\bar {H} \times \bar {W}}$, $\bar H$, $\bar W$ represents the height and width of the cropped patches, $\bar H = \bar W$. Each patch $\hat I_i$ is sent to the same iCNN to predict the pixel-wise labels $\widetilde z_i$:
	\begin{equation}
	\widetilde {z}_i =K_i({\hat I_i}),
	\end{equation}
where $\widetilde {z}_i \in \mathbb{R}^{\bar H \times \bar W}$. 

The previous steps were sufficient for training. As \ $\widetilde {z}_i$ \ is used for loss computation. The ground truth of \  $\widetilde {z}_i$ \  is cropped from the input labels by $\mathcal{T}_{\theta}$. 

The next steps are necessary to assemble the partial predictions and compute the $F_1$ score used in testing. This can be done by applying reverse grid transformer  $\mathcal{T}^{-1}_{\theta}$ to remap all the partial predictions back to original positions:
	\begin{equation}
	\widetilde{z}_i'  = \mathcal{T}^{-1}_{\theta}(\widetilde {z}_i),
	\end{equation}
where $\widetilde{z}_i' \in \mathbb{R}^{H \times W}$.

Finally, Softmax $S$ and channel-wise argmax $A$ are applied to get the final prediction $\hat z$:
\begin{equation}
  	  \hat z  = A(S(\widetilde{z}')).
\end{equation}
	
The pipeline process having been described. It is seen that there are two major modules, iCNN and STN, whose architectures are detailed in subsequent subsections.

\subsection{Interlinked CNN}
\label{subsubsec:icnn}

\begin{figure*}[t]
  \begin{center}
    \includegraphics[width=\linewidth]{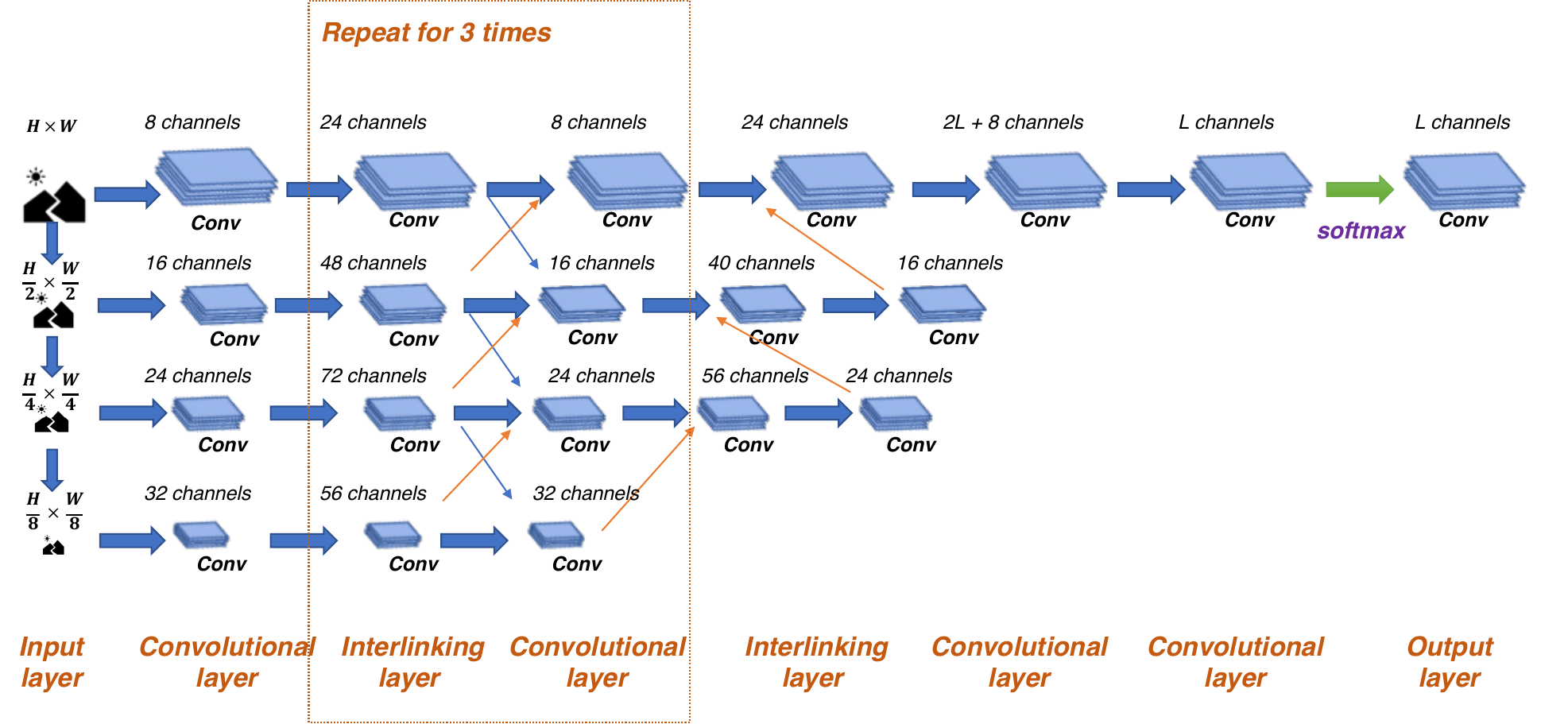}
  \end{center}
  \vspace{-1mm}
    \caption{The structure of iCNN. L: The number of label channels. Blue arrows: downsampling (\emph{max pooling}). Orange arrows: upsampling (\emph{nearest neighbor}). All convolutional layers share the same parameters: $kernel \ size = 3, \ stride = 1, \ padding = 1$.}  \label{fig:icnn}
\end{figure*}

iCNN is a CNN structure proposed by Zhou \etal \cite{zhou2015interlinked} for semantic segmentation. It is composed of four groups of fully convolutional networks with interlink structures, which can pass information from coarse to fine. The structure of iCNN model is illustrated in Fig. \ref{fig:icnn}. There are four CNNs that each use different sizes of filter. Each CNN uses only convolutional layers with no downsampling to maintain image size throughout the networks. In-between convolutional layers, there are interlinked layers. Each interlinked layer is applied on 2 layers at the same depth, with one layer having twice the length of the other (vertical neighbors on the figure). The smaller feature map is upsampled and concatenated with the feature map of the larger feature map similarly, the larger feature map is downsampled to be concatenated with the feature map of the smaller feature map.

In this work, we also use iCNN for both coarse and fine labeling stages. Without changing the original iCNN structure, we added Batch Normalization and ReLU activation to the network. Additionally, we also used larger pictures as inputs, from $64 \times 64$ to $ 128 \times 128 $.

\subsection{Spatial Transformer Network}
\label{subsubsec:STN}

The key to end-to-end training in our method is to connect the isolated gradient flow between the two training stage. To achieve this, a differentiable cropper needs to be implemented. Inspired by Tang \etal \cite{tang2019improving}, we use a modified version of STN \cite{jaderberg2015spatial} to perform positioning and region-based feature learning.
As described by Jaderberg \etal \cite{jaderberg2015spatial}, Spatial Transform Network (STN) is composed of Localization Net, Grid Generator and Sampler. In this paper, for the sake of simplicity, the combination of Grid Generator and Sampler corresponds to Grid Transformer.

\begin{figure}
    \includegraphics[width=\linewidth]{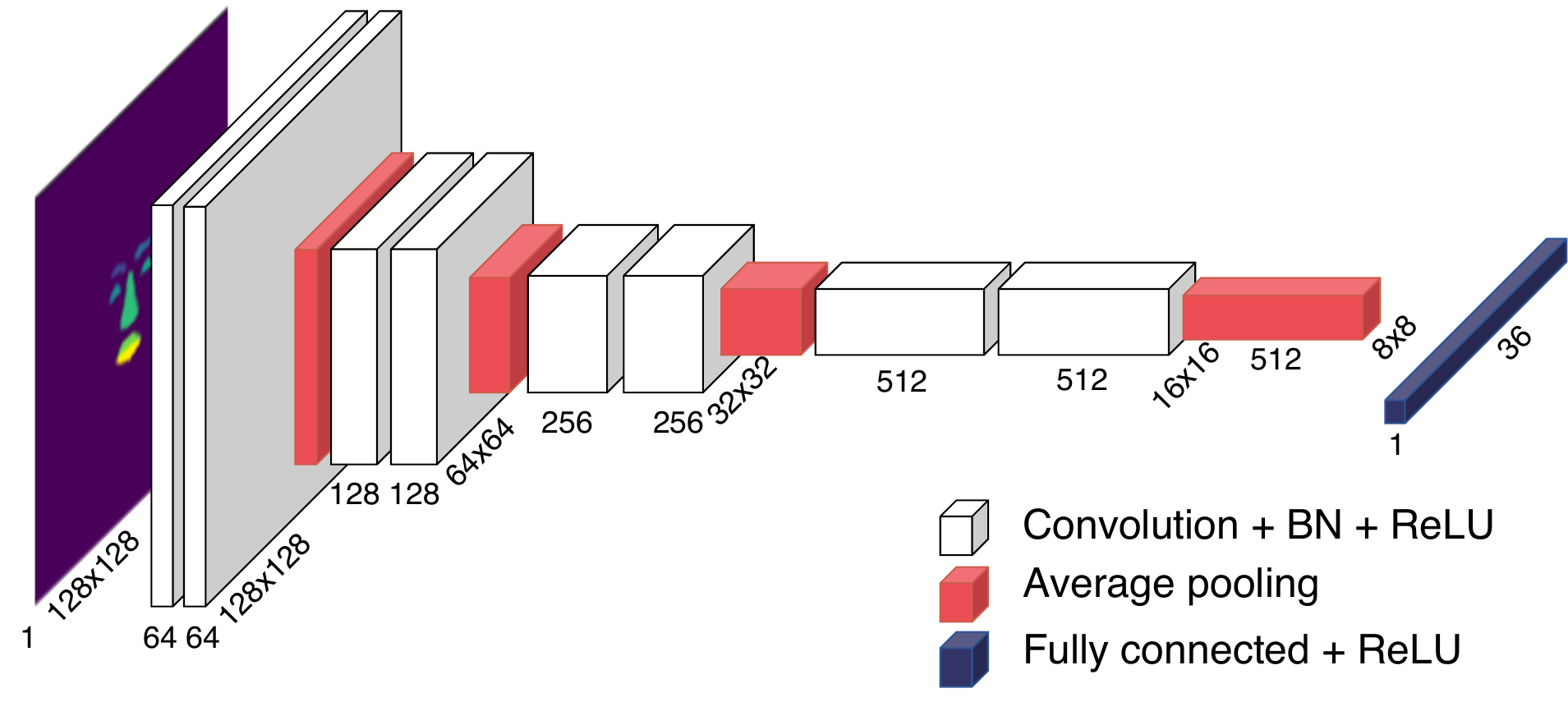}
    \caption{The structure of Localization Network in STN Module. 
This 9-layer network is a simplified version of VGG16. Each convolutional layer (white) includes a convolution, Batch Normalization and ReLU non-linear activation. After every two convolutional layer, an average pooling (red) is applied.  Finally, a fully connected layer is applied with ReLU activation (blue).  For all convolutional layers: $kernel \ size = 3,\ stride=padding=1$. For all pooling layers:$ kernel \ size = 3, \ stride = 2$.
}
\label{fig:ln}
\end{figure}
\paragraph{\textbf{Localization Network}} The 9-layer localization network we used is simplified from VGG16 \cite{simonyan2014very}. This network can be replaced with another convolutional neural network structure to obtain better performance. As Fig. \ref{fig:ln} shows, we first use 8 convolutional layers to perform feature extraction and map it to a $6 \times 2 \times 3$ transform matrix $\theta$ through a fully connected layer.

\paragraph{\textbf{Grid Transformer}}
The grid transformer samples the relevant parts of an image into a regular grid $ G$ of pixels $G_{i} = (x^{t}_{i},y^{t}_{i})$, forming an output feature map $V \in \mathbb{R}^{C \times \bar H \times \bar W}$, where $\bar H$ and $\bar W$ are the height and width of the grid, $C$ is the number of channels. We use 2D affine transformation for $\mathcal{T}_{\theta}$:
\begin{align}
\left(\begin{array}{c}{x_{i}^{s}} \\ {y_{i}^{s}}\end{array}\right)=\mathcal{T}_{\theta}\left(G_{i}\right)=\left[\begin{array}{ccc}{\theta_{11}} & {\theta_{12}} & {\theta_{13}} \\ {\theta_{21}} & {\theta_{22}} & {\theta_{23}}\end{array}\right]\left(\begin{array}{c}{x_{i}^{t}} \\ {y_{i}^{t}} \\ {1}\end{array}\right),
\end{align}
where $(x_{i}^S, y_{i}^S)$, $(x_{i}^t, y_{i}^t)$ are the source coordinates and target coordinates of the $i$-th pixel. In order for STN to do crop operations, we constrain $\theta$ as follows:
\begin{equation}
\theta=\left[\begin{array}{ccc}{s_{x}} & {0} & {t_{x}} \\ {0} & {s_{y}} & {t_{y}}\end{array}\right],
\end{equation}which allows cropping, translation, and isotropic scaling by varying $s$, $t_x$, and $t_y$. These parameters are predicted from the rough mask by localization network $L(x)$.

\subsection{Loss Function}
The average binary cross entropy loss has been used as a criterion for both coarse and fine segmentation:

\begin{equation}
\mathscr{L}_{cse}(p,t) = -\frac{1}{N} \Sigma_{i=1}^{N}  t_i \log({p}_i)+(1-t_{i})log(1-{p}_i),
\end{equation}
where $N$ is the number of parts, $p$ is the prediction and $t$ is the target ground truth.The loss function $\mathscr{L}_{F}$ of the entire system is defined as follows:

\begin{equation}
 \mathscr{L}_{F} = \frac{1}{N}\Sigma_i^N \mathscr{L}_{cse}(\widetilde {z}_i, \hat J_i),
\end{equation}
where $\widetilde {z}_i$ is parts predicton, $\hat J_i$ is the binary ground truth cropped from label $J$. $\mathscr{L}_{F}$ is used to optmizie the whole system.

\subsection{Implementation Details}
\label{subsec:train}

\subsubsection{Preprocessing}
\label{subsubsec:prepro}
 \paragraph{\textbf{Batch Processing}}
	In order to batch process images with different sizes, The input image is resized to the same size using bilinear interpolation and padding, respectively. The interpolated image is sent to the rough labeling network. The padded image is sent to the STN for cropping, because the padding operation can retain the original image information.
	
\paragraph{\textbf{Data Augmentation}}
	We perform data augmentation during data preprocessing, using a total of 4 random operations:
	
\begin{enumerate}[\hspace{2em} (a)]
	\item Random rotation,  range from  -15\degree to 15\degree 
	\item Random shift, horizontal shift is randomly sampled in the range $(-0.2w, 0.2w)$ and vertical shift is randomly sampled in the range $(-0.2h, 0.2h)$, where $w$, $h$ represents width and height of an image.
	\item Random scale, the scale is randomly sampled from the range $[0.2, 1.2]$.
	\item Gaussian random noise, this operation is not performed on labels.
\end{enumerate}

Each image is augmented into 5 images and each augmentation takes from 0 to 4 of the operations at random. An image $I$ will be augmented into $I_{i}$ with $i \in \{0,1,2,3,4\}$ and each $I_{i}$ will have $i$ of the 4 operations applied randomly. For instance, an image $I$ will be augmented into 4 different images by applying sets of operations (a), (b, d), (a, c, d), and (a, b, c, d).

\subsubsection{Training}
\begin{figure*}
\centering
    \includegraphics[width=0.85\linewidth]{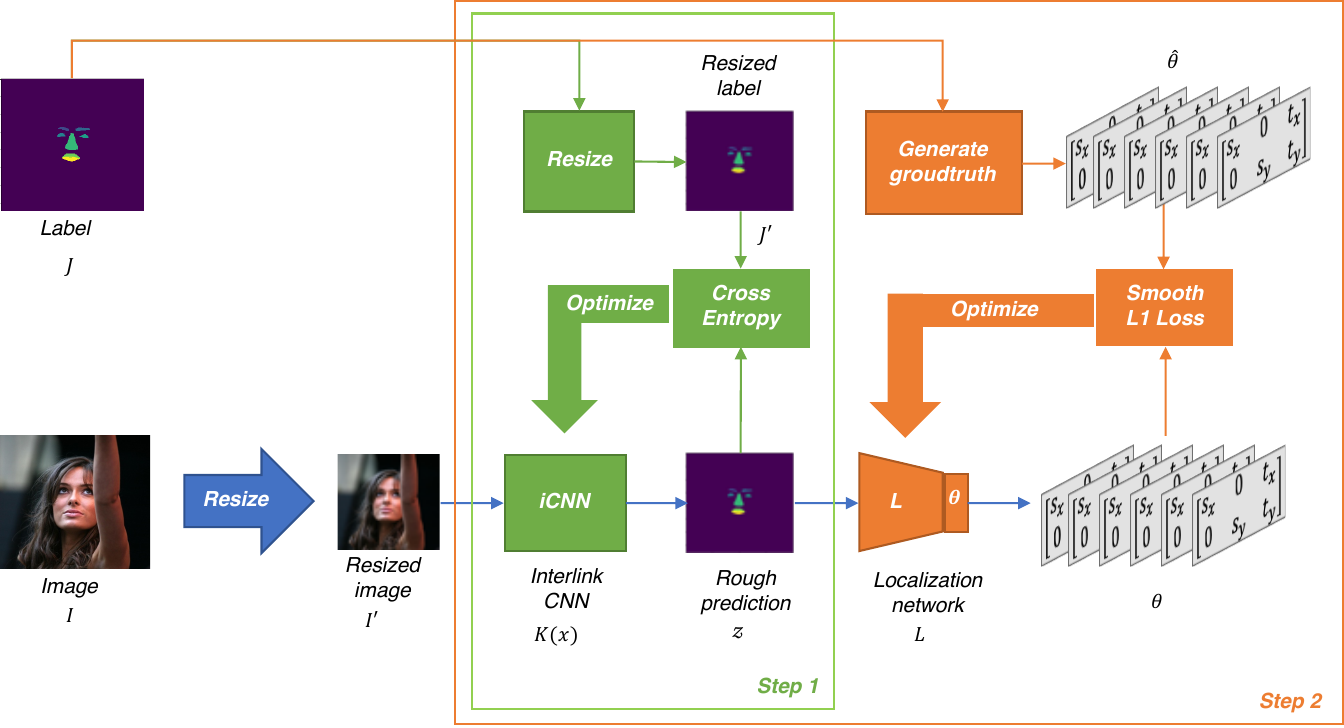}
    \caption{Pre-training Pipeline. }
    \label{fig: pretrain}
\end{figure*}
We divide the training process of the entire system into pre-training and end-to-end training.

 \paragraph{\textbf{Pre-training}} For better results, we pre-trained the system. There are two modules that need to be pre-trained, iCNN $ K $ for coarse segmentation, and localization network $ L $ for parts localization. We perform the pre-training operation in two steps: first $ K $ is pre-trained, then using the parameters of $K $ to pre-train $L$.  As shown in Fig. \ref{fig: pretrain}, the input of $ K $ is the resized image $ I'$, and the output is the rough prediction $ z $. Its optimization goal is \emph{CrossEntropy} loss $\mathscr{L}_{R}$ between $ z$ and resized label $J'$:
 \begin{equation}
 	\mathscr{L}_{R} = \mathscr{L}_{cse}(z, J'),
 \end{equation} 
where $J'$ is resized label. 

The input of $L$ is $z$, and the output is transformer matrix $\theta$, its optimization goal is \emph{Smooth L1} loss between $\theta$ and $\hat \theta$, the formulation of \emph{Smooth L1} is:
\begin{equation}
\mathscr{L}_{SmoothL1Loss}(\theta,\hat \theta)=\frac{1}{N}\Sigma_{i}^{N}u_i,
\end{equation}
where $u_i$ is given by:
\begin{equation}
	u_{i}=\left\{\begin{array}{ll}{0.5\left(\theta_{i}-\hat {\theta_{i}}\right)^{2},} & {\text { if }\left|\theta_{i}-\hat{\theta}_{i}\right|<1} \\ {\left|\theta_{i}-\hat{\theta_{i}}\right|-0.5,} & {\text { otherwise. }}\end{array}\right.
\end{equation}
It uses a squared term if the absolute element-wise error falls below $1$ and an $L1$ term otherwise. $\hat \theta$ is the groundtruth of $\theta$, which can be generated from original label $J$. The details of the generation of $\hat \theta$ are as follows. 
 
 Given binary label $J_i \in \mathbb{Z}^{H \times W}$, the central coordinates $(x, y)$ of parts can be calculated. And with the cropping window size fixed to $ \bar H \times \bar W $, $\hat \theta$ can be calculated by equation (\ref{equ:thetag}).
 \begin{equation}
 \label{equ:thetag}
\hat \theta = \left[\begin{array}{ccc}{S_x} & {0} & {t_{x}} \\ {0} & {S_y} & {t_{y}}\end{array}\right]=\left[\begin{array}{ccc}{\frac{\bar W}{W}} & {0} & {-1+\frac{2x}{W}} \\ {0} & {\frac{\bar H}{H}} & {-1+\frac{2y}{H}}\end{array}\right].
 \end{equation}

\paragraph{\textbf{End-to-end training}} With pre-trained parameters loaded, we perform end-to-end training on the whole framework:
\begin {equation}
\label{equ:Fx}
\widetilde z =F(I),
\end {equation}
where $F$ represents the whole proposed framework. The optimization goal of the system is the cross-entropy loss between parts prediction $ \widetilde z $ and partial labels $\hat J$ cropped from $J$. At this stage, the learning rate of the pre-trained networks is lower than that of other networks.

\paragraph{\textbf{Optimization}} We updated the network parameters using the \emph{Adam} algorithm \cite{kingma2014adam}. We halved the learning rate every 5 epochs using a scheduler.

\begin{table}[htbp]
\centering
\caption{Hyperparameter settings. Lr1: learning rate of rough labeling iCNN, Lr2: learning rate of Localization Network, Lr3: learning rate of partial labeling iCNNs. GPU Nums: numbers of NVIDIA GTX 1080Ti GPU.}
 \label{tab:settings}       
\begin{tabular}{C{1.5cm}|C{0.8cm}C{0.8cm}C{0.8cm}C{0.8cm}C{0.8cm}C{0.8cm}}
  \toprule
  \multicolumn{7}{c}{HELEN} \\
  \hline
  Stage & Lr1 & Lr2 & Lr3 & Batch Size & Epochs & GPU Nums\\
  \hline
Pre-training & 0.0023 & 0.0017 & - & 32 & 25 &1x\\
  \hline
End-to-End  & $10^{-4}$ & $10^{-4}$ & $10^{-2}$ & 32 & 25 &1x\\
  \hline
       \multicolumn{7}{c}{CelebAMask-HQ} \\
         \hline
    Stage & Lr1 & Lr2 & Lr3 & Batch Size & Epochs & GPU Nums\\
  \hline
Pre-training & $10^{-3}$ & $10^{-3}$ & - & 64 & 25 &4x\\
  \hline
End-to-End  & $10^{-6}$ & $10^{-6}$ & $10^{-3}$ & 64 & 25 & 4x\\
  \bottomrule
 \end{tabular}
\end{table}

\section{Experiments}
\label{sec:exp}
\subsection{Datasets and Evaluation Metric}
\label{subsec:4.1}
\paragraph{\textbf{Datasets}} To have comparable results with \cite{zhou2015interlinked}, we performed our experiments on the HELEN Dataset \cite{smith2013exemplar}. The HELEN dataset contains $2330$ images. Each image is annotated with binary masks labeled by $11$ categories: background, face skin, eyes (left, right), eyebrows (left, right), nose, upper-lip, lower-lip, inner mouth and hair. Like \cite{zhou2015interlinked,liu2015multi,yamashita2015cost,wei2017learning,lin2019face}, we split the dataset into training, validation, and test set with respectively $2000$, $230$, and $100$ samples.

We also applied our model on CelebAMask-HQ dataset \cite{celebamaskhq}. CelebAMask-HQ is a large-scale facial image dataset. It consists of $30,000$ high-resolution face images selected from CelebA\cite{celeba}. Each picture is semantically labelled with labels corresponding to specific facial elements. These semantic tags are divided into $19$ categories, and these categories contain both facial elements and accessory items including: skin, nose, eyes, eyebrows, ears, mouth, lips, hair, hats, glasses, ears, necklaces, necks, and clothes. In our experiments, we only used the semantic labels shared with those from the HELEN dataset, and we divide the dataset into a training set, a test set, and a validation set according to the ratio of $6 : 2 : 2$. This corresponds to $18,000 $, $ 6000 $, and $ 6000 $ images, respectively.

\paragraph{\textbf{Evaluation metric}} 
This paper uses the $ F_1 $ score as the evaluation metric. The $F_1$ score is the harmonic mean of the precision and recall, and reaches its best value at 1 (perfect precision and recall) and worst at 0. The metric is formulated as follows:
\begin{align}
P &=\frac{TP}{TP+FP}, \\ 
R &=\frac{TP}{TP+FN}, \\
F_{1} &=\frac{2 \times P \times R}{P+R},
\end{align}
where TP denotes True Positive predictions, FP False Positive predictions, and FN False Negative predictions.

\paragraph{\textbf{Hyperparameters and Efficiency}}

Table \ref{tab:settings} shows the detailed hyperparameters. We trained the proposed model in two stages: pre-training and end-to-end. In particular, the pre-trained parameters obtained from the previous stage are loaded before end-to-end training. For HELEN dataset, both training and inference are performed on a single NVIDIA GTX1080Ti GPU, while for CelebMaskA dataset these operations are executed on four NVIDIA GTX1080Ti GPUs instead, each with a batch size of 16. 

The proposed model have good efficiency. For inference efficiency, the baseline model runs at 86 ms per face, and the proposed model runs at 80 ms per face. For training efficiency on HELEN, the pre-training stage takes 0.5 hours, and the end-to-end stage takes 2.2 hours. For training efficiency on CelebMaskA, the pre-training stage and end-to-end stage requires 1.6 hours and 10.2 hours, respectively.

\paragraph{\textbf{ Hybrid training strategy on CelebMaskA: Training on HELEN, Fine-tuning on CelebMaskA}} Due to the large amount of data in CelebMaskA, direct training requires more computing resources and longer training time. In practice, we found out that by using 2000 images from CelebMaskA to fine-tune a model already trained on the Helen Dataset, the performances were similar. This method only takes 2.1 hours on a single GPU, saving a great amount of training time. This demonstrates the generalization of our model.

\subsection{Comparison with the Baseline Model on HELEN}
\begin{table*}[htbp]
\centering
\caption{$F_1$ scores of different models on HELEN.}
 \label{tab:baseline}       
\begin{tabular}{C{3cm}|C{1cm}C{1cm}C{1cm}C{1cm}C{1cm}C{1cm}C{1cm}C{1cm}C{1cm}}
  \toprule
  Methods & eyes &brows & nose & In-mouth & Upper-lip & Lower-lip & mouth & skin & overall \\
  \hline
iCNN (Baseline) & 0.778 & $\mathbf{0.863}$ & 0.920 & 0.777 & $\mathbf{0.824}$ & 0.808 & 0.889 & - & 0.845 \\
iCNN (Our implem.)  & 0.863 & 0.790 & 0.936 & 0.812 & 0.772 & 0.830 & 0.908 & - & 0.865 \\
STN-iCNN\textsuperscript{*}  & 0.891 & 0.845 & 0.956 & 0.853 & 0.792 & 0.834 & 0.920  & - & 0.893 \\
STN-iCNN   & $\mathbf{0.895}$ & $0.848$ & $\mathbf{0.963}$ & $\mathbf{0.856}$ & $\mathbf{0.824}$ & $\mathbf{0.866}$ & $\mathbf{0.928}$ & - & $\mathbf{0.910}$ \\
  \bottomrule
  \multicolumn{2}{c}{* denotes training without end-to-end.}
 \end{tabular}
\end{table*}
We used the results in \cite{zhou2015interlinked} as the baseline, and compared it with the results of reimplemented iCNN and the proposed STN-iCNN on the HELEN dataset. The comparison results are shown in Table \ref{tab:baseline}, where STN-iCNN* represents the results of STN-iCNN before end-to-end training. 

As shown in the Table \ref{tab:baseline}, the model's results have improved significantly even before end-to-end training. This is because the Localization Network in STN has a deep CNN layer, so it can learn the context relationship of the semantic parts from the rough mask. In the case where the rough mask is not complete, the accurate transformer matrix can still be predicted. Therefore, STN is able to crop more accurately than the original cropper, which improves the overall performance. 

\begin{figure}
\includegraphics[width=0.48\textwidth]{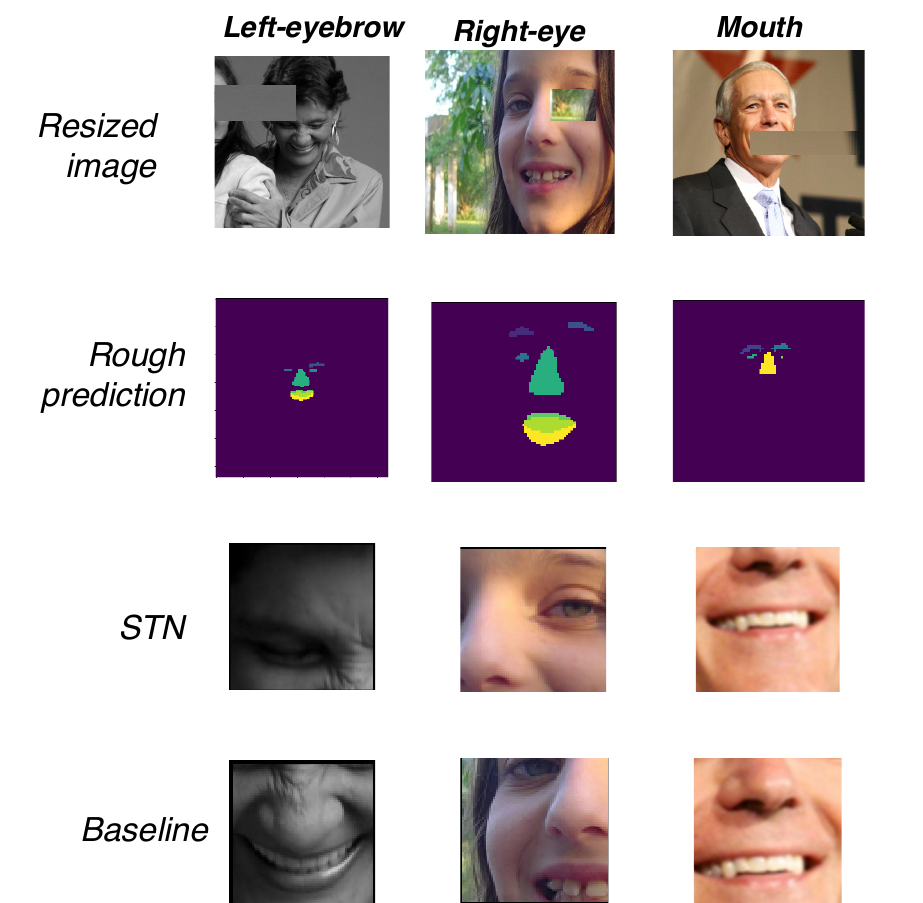}
\caption{Comparison of two cropping methods on HELEN, the Baseline method and STN method.}
\label{fig:chal}
\end{figure}

 As shown in Fig. \ref{fig:chal}, we performed a comparison experiment of two different cropping methods on the HELEN dataset. In this experiment, we selected some images and randomly cover part of their facial components (such as left eyebrows, right eye, mouth, etc.) with their background information. We then sent the images to the rough labeling model to get the incomplete rough segmentation results, see row 2 in Fig. \ref{fig:chal}. Based on the rough results, we used the baseline method and STN method to crop unresized images and compare their cropping results. The experimental results are shown in the last two rows of Fig. \ref{fig:chal}. The results show that the STN method can work normally even if the rough mask is partially missing.

\begin{table*}[htbp]
\centering
\caption{$F_1$ scores of different models on CelebAMask-HQ}
 \label{tab:celeb}       
\begin{tabular}{C{3cm}|C{1cm}C{1cm}C{1cm}C{1cm}C{1cm}C{1cm}C{1cm}C{1cm}C{1cm}}
  \toprule
  Methods & eyes &brows & nose & In-mouth & Upper-lip & Lower-lip & mouth & skin & overall \\
  \hline
iCNN((Our implem.) & 0.883 & 0.848 & 0.910 & 0.896 & 0.712 & 0.763 & 0.849 & - & 0.879 \\
STN-iCNN\textsuperscript{*}  & 0.866 & 0.834 & 0.937 & 0.906 & $\mathbf{0.878}$ & $\mathbf{0.893}$ & $\mathbf{0.948}$  & - & 0.915 \\	
STN-iCNN   & $\mathbf{0.900}$ & $\mathbf{0.851}$ & $\mathbf{0.943}$ & $\mathbf{0.917}$ & 0.877 & $\mathbf{0.893}$ & $\mathbf{0.948}$ & - & $\mathbf{0.924}$ \\
  \bottomrule
  \multicolumn{2}{c}{* denotes training without end-to-end.}
 \end{tabular}
\end{table*}
\subsection{Sensitivity of Hyperparameters}
\begin{figure}
\includegraphics[width=0.48\textwidth]{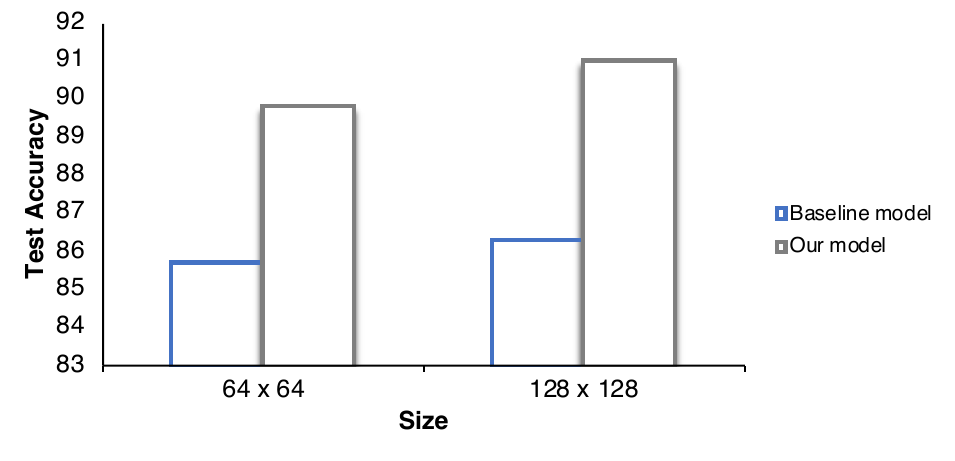}
\caption{Comparison of the effect of $b$ on baseline model and our model. This experiment was conducted on the HELEN dataset.}
\label{fig: resize}
\end{figure}

\paragraph{\textbf{The size after resizing}} The baseline input size for model K is $64 \times 64$, taking into account the smaller size of eyes and eyebrows features, the input size was changed to $128 \times 128$ in our model. This change had limited effect on the baseline method, but improved our approach significantly, see Fig. \ref{fig: resize}.

\begin{figure}[htbp]
\centering
\includegraphics[width=0.48\textwidth]{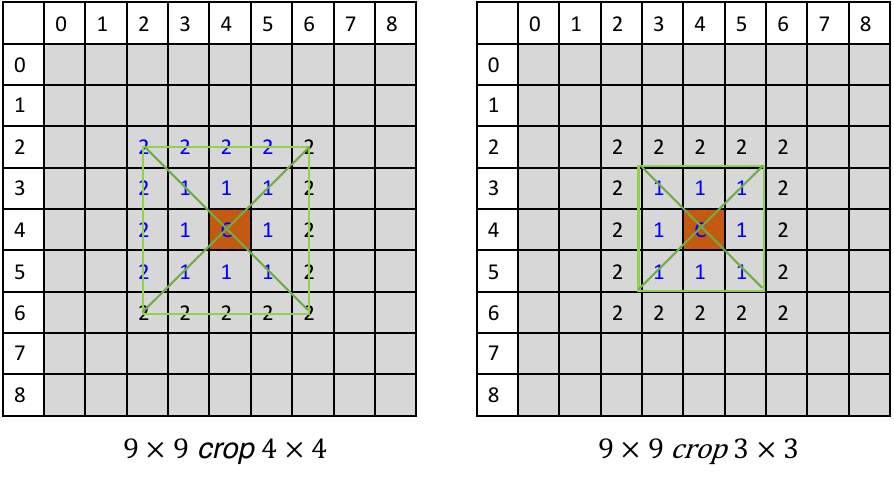}
\caption{Comparison between two cropping methods on the even size patch and on the odd size. The blue one is the baseline cropper, and the green one is STN method. }
\label{fig: grid}
\end{figure}

\begin{figure}\centering
\includegraphics[width=0.48 \textwidth]{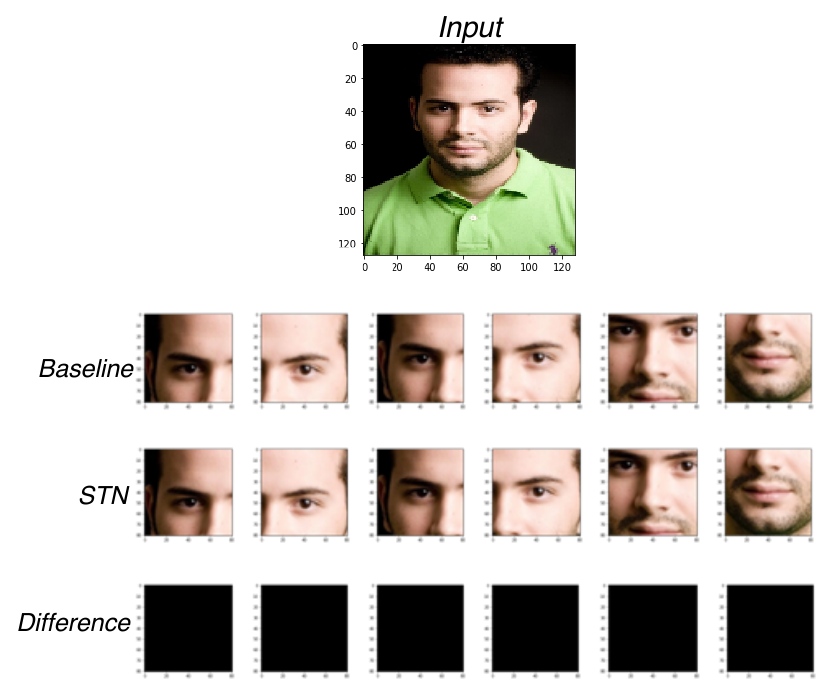}
\caption{Parts cropped by two different cropping methods.The first row is a randomly selected facial image from the HELEN dataset, and the second row is parts cropped by the baseline cropper. The third row is parts cropped by STN. The last row shows the pixel-wise difference between the two pictures, which is all zero.}
\label{fig: two_crop}
\end{figure}

\paragraph{\textbf{The size of cropped patches}}The size of the cropped patch should be odd rather than even. This is to ensure the integer grid coordinates during grid sampling so that the grid transformer $ \mathcal{T}_{\hat \theta}(G) $ is equal to the cropper of baseline method in the crop operation.  As shown in Fig. \ref {fig: grid}, it can be seen that when an integer gird cannot be found, STN performs bilinear interpolation while the baseline cropper makes one-pixel offset, thus leads to an unequal result. For the HELEN dataset, we set $ \bar H = \bar W = 81 $, and for the CelebMaskA dataset, we set $ \bar H = \bar W = 127 $. The comparison results of the two cropping methods on the HELEN dataset are shown in Fig. \ref {fig: two_crop}.

\subsection{Comparison with State-of-the-art Methods on HELEN}

\begin{table*}
\centering
 \caption{Comparison with State-of-the-art Methods on HELEN}
 \label{tab:test}
\begin{tabular}{c|C{1cm}C{1cm}C{1cm}C{1cm}C{1cm}C{1cm}C{1cm}C{1cm}C{1cm}}
  \toprule
  Methods & eyes &brows & nose & In-mouth & Upper-lip & Lower-lip & mouth & skin &overall \\
  \hline
Smith \etal \cite{smith2013exemplar} & 0.785 & 0.722 & 0.922 & 0.713 & 0.651 & 0.700 & 0.857 & 0.882 &0.804 \\
Zhou \etal \cite{zhou2015interlinked} & 0.778 & $\mathbf{0.863}$ & 0.920 & 0.777 & $\mathbf{0.824}$ & 0.808 & 0.889 & - & 0.845 \\
Liu \etal \cite{liu2015multi} &0.768&
 0.713 & 0.909 & 0.808 & 0.623 & 0.694 & 0.841 & 0.910 & 0.847  \\
Liu \etal \cite{liu2017face} & 0.868 & 0.770 & 0.930 & 0.792 & 0.743 & 0.817 & 0.891 & 0.921 & 0.886 \\
Wei \etal \cite{wei2017learning} & 0.847 & 0.786 & 0.937 & - & - & - & 0.915 & 0.915 & 0.902 \\
\hline
iCNN (Our implem.) & $0.863$ & $0.790$ & $0.936$ & $0.812$ & $0.772$ & $0.830$ & $0.908$ & - & $0.865$ \\
STN-iCNN   & $\mathbf{0.895}$ &  $0.848$ & $\mathbf{0.963}$ & $\mathbf{0.856}$ & $\mathbf{0.824}$ & $\mathbf{0.866}$ & $\mathbf{0.928}$ & - & $\mathbf{0.910}$ \\
  \bottomrule
 \end{tabular}
\end{table*}

After selecting appropriate hyperparameters, we completed the end-to-end training of STN-iCNN proposed in this paper and compared its test results with state-of-the-art on the HELEN dataset. As can be seen from Table \ref{tab:test}, the performance of the original iCNN model has been greatly improved by the proposed STN-iCNN. It is worth mentioning that because our model cannot handle hair, and we cannot determine the effect of hair on the overall score of our model, so we did not compare it with the results of \cite{lin2019face}.

\subsection{Results on CelebAMask-HQ}

Table \ref{tab:celeb} shows the comparison of the baseline model and our proposed model for $ F_1 $ score. The results on CelebAMask-HQ once again proved the effectiveness of the proposed method and the end-to-end training, indicating that our model has a certain generalization ability.

\section{Conclusion}
\label{sec:conc}
We introduced the STN-iCNN, an end-to-end framework for face parsing, which is a non-trivial extension of the two-stage face parsing pipeline presented in \cite{zhou2015interlinked}. By adding STN to the original pipeline, we provide a trainable connection between the two isolated stage of the original method, and successfully achieve end-to-end training. The end-to-end training can help two labeling stages optimize towards a common goal and improve each other, so the trained model can achieve better results. Moreover, the addition of STN also greatly improves the accuracy of facial component positioning and cropping, which is also important for overall accuracy. 

Experimental results show that a simple modification of the baseline model significantly improves the parsing score. This proves that end-to-end training helps to improve the performance of the model, thereby facilitating future face parsing studies.

STN-iCNN can be regarded as a region-based end-to-end semantic segmentation method that does not require extra position annotations. In spite of this, it has limitation on handling hair and skin, and that is why it ignores the face and skin just like most existing models. One possible solution is to adopt separate mask branches and appropriate preprocessing methods, which is beyond the scope of this paper.

In future studies, we will solve the problem of hair segmentation and further improve the performance. Besides, extending the proposed framework to general semantic segmentation task is also a valuable direction.

\noindent\textbf{Acknowledgements} \
This work was supported by the National Natural Science Foundation of China under Grant Nos. U19B2034, 51975057, 601836014. The first author would like to thank Haoyu Liang, Aminul Huq for providing useful suggestions.

{\small
\bibliographystyle{ieee_fullname}
\bibliography{iccv}
}

\end{document}